# Detecting Toxic Language: Ontology and BERT-based Approaches for Bulgarian Text

Melania Berbatova and Tsvetoslav Vasev

Faculty of Mathematics and Informatics, Sofia University "St. Kliment Ohridski"

`{msberbatova, cvasev}@fmi.uni-sofia.bg`

*Warning: this paper contains content that may be offensive or otherwise inappropriate.*

**Abstract.** Toxic content detection in online communication remains a significant challenge, with current solutions often inadvertently blocking valuable information, including medical terms and text related to minority groups. This paper presents a more nuanced approach to identifying toxicity in Bulgarian text while preserving access to essential information.

The research explores two distinct methodologies for detecting toxic content. The developed methodologies have potential applications across diverse online platforms and content moderation systems. First, we propose an ontology that models the potentially toxic words in Bulgarian language. Then, we compose a dataset that comprises 4,384 manually annotated sentences from Bulgarian online forums across four categories: toxic language, medical terminology, non-toxic language, and terms related to minority communities. We then train a BERT-based model for toxic language classification, which reaches a 0.89 F1 macro score. The trained model is directly applicable in a real environment and can be integrated as a component of toxic content detection systems.

## 1 Introduction

Modern toxic content filters, such as those used by social networks like Facebook, often block content that is related to information about reproductive health and sexually transmitted diseases. Access to such information is vital for a large number of people who need it to address health problems in these areas. Often the reason for the misclassifications is that biological terms related to the reproductive systems are part of the words marked as toxic, as in the Toxicity-200 dataset [6]. Similar problems arise with content related to minority groups or with the use of words with multiple meanings. These inefficiencies can lead to unfair blockages and limit free access to important information. Improving toxic content filters brings significant value to both users and online platforms (social networks, forums, etc.) by creating a better and more welcoming environment while reducing the risk of wrongly blocking important information.

In this paper, we suggest a toxic language ontology that models concepts such as toxic language, medical terminology, non-toxic language, words and expressions related to minority groups, as well as different internet domains such as family-friendly pages and internet forums. We then demonstrate the implementation of toxic content classifiers for Bulgarian language using two approaches: ontology-based and BERT-based. Finally, we discuss how these approaches can be accommodated in the domain of large language models.

The suggested ontology would be useful to media and linguistics researchers who analyze online content, as it formally conceptualizes toxic language and its use in a variety of contexts. The implemented classifiers recognize toxic language in different contexts due to the presence of the additional categories: medical terminology, non-toxic language, words and expressions related to minority communities. Given this, they can be used by social platforms, online forums, content filters of text generated by large language models (LLMs), and others. The use of such systems would allow better protection of users from unwanted and offensive messages while maintaining access to important information. The code produced for the research is publicly available on GitHub[1].

---

[1] https://github.com/TsvetoslavVasev/toxic-language-classification

## 2    Related work

### 2.1    Challenges of Toxic Language Detection

Toxic speech in the cyberspace is on the rise and therefore it is necessary to build systems to automatically detect or filter such content. Various supervised machine learning approaches need manually annotated toxic speech corpora, but the task of identifying toxic content often hinders the annotators' work due to the subjectivity of ratings, context, cultural differences, and the emotional impact of prolonged exposure to such language [16].

The use of toxic speech has a wide range, affecting different areas of our lives, including communication in our workplace. Models trained on existing hyper-toxic datasets and microaggression datasets do not detect veiled toxic text. To some extent, context helps BERT-based models detect less harsh toxic sentences, but efforts are still needed in this area [4].

There are structural differences between toxic and non-toxic comments in different languages [11] and this specificity should be taken into account in studies to detect such content. A major challenge in the development of toxic language recognition models in Bulgarian is the lack of sufficient annotated data, as the creation of a high-quality corpus will require significant resources and time. The fact that numerous words have multiple meanings depending on the context makes the task of identifying toxic language even more difficult. Contextual models such as BERT and ELMo partially address this problem, but there is still room for improvement. The detection of toxic language may affect specific demographic groups and therefore the identification of such content should be done carefully in order not to violate human rights and freedom of speech. In this regard, it is necessary to establish clear criteria for toxicity and avoid excessive censorship.

### 2.2    Current State of Toxic Language Classification

The automatic identification of abusive language is a long-studied subject. Solutions have been varying from linear and tree-based classifiers with manually engineered features [10], neural architectures like recurrent neural networks (RNN) and convolutional neural networks (CNN) [23], fine-tuning pre-trained language models [5], to more recently using large language models (LLMs) and prompt engineering [3,26].

While the zero-shot of LLMs is successfully incorporated for a variety of other classification tasks, research shows that for toxicity detection, their performance falls short compared to transformer models trained on in-domain data. Therefore, fine-tuned transformer models, such as BERT, RoBERTa and XLM and their fine-tuned variations remain central to state-of-the-art toxicity detection. Recent studies report such models achieving between 0.90 and 0.95 F1 scores for the binary toxic classification tasks in English [17], and between 0.80 and 0.82 in a multilingual setting [26].

### 2.3    Toxic Language Resources

There are numerous language resources such as lexicons, dictionaries, ontologies and datasets related to the topic of toxic language [1,2,6,14,22]. An example of such a lexicon is HurtLex [1], which is multilingual, and contains offensive and aggressive words as well as those expressing hate. The words are divided into 17 categories, including negative stereotypes; physical disability; cognitive disability; moral and behavioural defects; male genitalia; female genitalia; prostitution; homosexuality; derogatory words; crimes or words associated with immoral behaviour, etc. The lexicon contains annotated words and phrases in 50 languages, including Bulgarian.

In addition to HurtLex, several other datasets and lexicons have been developed for toxic language detection. One of them is the Flores-200 Toxicity Dataset [6], a comprehensive multilingual resource curated by META (formerly Facebook). This lexicon encompasses 200 languages and categorizes words deemed toxic, including vulgar language, hate speech, pornographic terms, and body parts, thus providing a broad foundation for harmful content detection.

Another significant contribution to the field is the Bulgarian Toxicity Dataset [20]. This dataset comprises a substantial collection of annotated sentences, categorized into hate speech, vulgarity, and other forms of toxic language. Building upon this foundation, we have created an enhanced dataset, incorporating additional data from Bulgarian online forums and manually annotating it into four distinct categories.

The utility of these datasets is largely contingent on their comprehensiveness and the specificity of their annotations. While HurtLex offers a robust framework for identifying various forms of toxic language across multiple linguistic contexts, it may benefit from additional context-specific annotations to enhance its accuracy in diverse cultural settings. The Flores-200 Toxicity Lexicon, despite its extensive coverage, primarily focuses on overtly toxic terms, potentially limiting its efficacy in detecting more nuanced forms of toxicity or context-dependent meanings.

The Bulgarian Toxicity Dataset and our newly developed dataset represent valuable resources for the development and evaluation of toxic language detection models specific to the Bulgarian language. The manual annotation process ensures high accuracy in sentence categorization, although it is important to acknowledge the potential for inherent subjectivity among human annotators. These datasets are particularly valuable for training models to detect toxic language within the Bulgarian context, rendering them highly relevant for both research endeavors and practical applications in content moderation.

### 2.4   Toxic Language Detection for Bulgarian Language

There are two main published studies in the field of toxic language detection in the Bulgarian language. In [8] the automatic recognition of toxic language in Bulgarian news articles is considered. The authors create a corpus of data from publications in the Bulgarian language manually annotated in eight categories (fake news, sensationalism, hate speech, conspiracies, anti-democratic, authoritarian, slander, delusion). Due to the small size of the corpus, for each type of features, the team trains a separate model, finally combining the models into a common "meta-classifier".

In [20] Ralev and Pfeffer develop different classifiers for the automatic recognition of hate speech in Bulgarian. The models were trained on a corpus of data annotated by the authors, extracted from Bulgarian internet forums. In their experiments, the authors determined the classifier based on the support vector method as the most effective.

The existing research on toxic language detection in Bulgarian has shown promising results, albeit with certain limitations. The study on "Detecting Toxicity in News Articles" [8] employs a meta-classifier approach, which enhances accuracy by capitalizing on the strengths of individual models trained on specific features. However, the restricted size of their corpus constrains overall performance, and the models exhibit difficulties in generalizing to more di- verse datasets. In contrast, the "Hate Speech Classification in Bulgarian" study demonstrates that their SVM-based classifier outperforms other tested methods, achieving the highest accuracy. Nevertheless, this approach faces challenges, including a dependency on feature engineering and difficulty in capturing the intricate linguistic patterns inherent in toxic language, particularly when confronted with nuanced or context-dependent expressions.

The strengths of the previous studies lie in their successful establishment of foundational models for detecting toxic language in Bulgarian and their innovative use of SVMs and meta-classifiers to improve accuracy. However, these studies were limited by small, specialized datasets and reliance on traditional machine learning models that require extensive feature engineering and struggle with context-dependent language.

## 3     The Toxic Language Ontology

We address the problem of nuanced toxic language detection by creating an ontology data model, which we name **ToxicOntoBg**[2]. We base our model on the Flores Toxicity-200 dataset [6], which is publicly available on GitHub[3]. We selected this dataset as the cornerstone for our ontology because of its precision in identifying unequivocally toxic language. While alternative datasets, such as HurtLex, offer an extensive array of offensive and aggressive terminology, they frequently encompass words that lack universal recognition as toxic or exhibit context-dependent interpretations. In contrast, the Flores Toxicity-200 dataset focuses specifically on words and phrases that are indisputably harmful, vulgar, or associated with hate speech, thereby providing a more reliable foundation for constructing an ontology aimed at recognizing toxic content.

### 3.1     Data Processing

The Flores 200 Toxicity Dataset contains a set of words and phrases in 200 languages classified as toxic by translators. The words and phrases included fall into one or more of the following categories:

1. Frequently used profanities
2. Frequently used insults and hate speech terms, or language used to bully, denigrate, or demean
3. Pornographic terms
4. Terms for body parts associated with sexual activity

The Bulgarian version of the lexicon contains 299 toxic, obscene or offensive words, covering a wide range of potentially inappropriate language. This resource serves as a starting point for identifying toxic content on which additional methods and techniques can be built. However, the lexicon also has limitations. It includes a fixed set of words and cannot cover all possible variations and combinations of toxic language. In addition, the lexicon does not take into account the context in which the words are used, which can be crucial for their correct interpretation. To model the different cases in which these words can be used, we distinguish the following 4 categories: toxic language, medical terminology, minority group language, and non-toxic language.

- *Toxic language* is defined as "rude, disrespectful, or unreasonable language that is likely to make someone leave a discussion." [12]
- *Medical terminology* is "language that is used to describe anatomical structures, processes, conditions, medical procedures, and treatments." [18]
- *Minority group* language refers to "subgroups of the population with unique social, religious, ethnic, racial, and/or other characteristics that differ from those of a majority group." [19]
- *Non-toxic language* refers to communication that is free from harmful, abusive or offensive content.

---

[2] https://huggingface.co/datasets/sofia-uni/toxic-onto-bg

[3] https://github.com/facebookresearch/flores/blob/main/toxicity

- A fair amount of the words in the corpus have multiple meanings, including toxic and non-toxic ones. The class "*Ambiguous*" consists of the words that have both toxic and neutral meanings.[4]

The process of manually annotating each Flores 200 Toxicity word into one or more of the categories mentioned above is essential for further modelling concepts such as toxic and non-toxic language, medical terminology, and terms associated with minority communities.

It should be noted that in the annotation process, each word was categorized manually by the authors. Although efforts were made to adhere to clear definitions and criteria for each category, it is necessary to emphasize the subjective nature of the assessments made.

When analysing the distribution (tables 1 and 2), it is noticeable that every word is classified as toxic. The majority of the words have only a strictly toxic meaning, but there are also a number of words falling into the other categories. Examples of words that fall into more than one category are:

- The word "*печка*" in Bulgarian means "stove" in everyday usage, but it can also be used as an insult to someone with a darker skin colour. It is an instance of both "ToxicLanguage" and "NonToxicLanguage" classes.
- The word "*седалище*" could mean "headquarters", but also "seat". It is an instance of "NonToxicLanguage" and "MedicalTerminology".

**Table 1.** Number and percentage of the words in the main ontology classes

| Toxic | MedicalTermnology | NonToxic | MinorityGroup |
|---|---|---|---|
| 299 (100%) | 32 (10.7%) | 32 (10.7%) | 12 (4%) |

**Table 2.** Co-occurrence matrix of the class labels

|  | Toxic | MedicalTerminology | NonToxic | MinorityGroup |
|---|---|---|---|---|
| **Toxic** | 299 | 32 | 32 | 12 |
| **MedicalTerminology** | 32 | 32 | 6 | 1 |
| **NonToxic** | 32 | 6 | 32 | 2 |
| **MinorityGroup** | 12 | 1 | 2 | 12 |

### 3.2 Ontology Design

The creation of an ontology allows for formalizing the knowledge contained in the lexicon and modelling the relationships between the different concepts. This approach will contribute to a more accurate representation and handling of the information needed for effective recognition and classification of terms in a Bulgarian text. The ontology is created based on the manually annotated lexicon using OWLReady2.

---

[4] As all the words in the current data corpus are annotated as potentially toxic, there is a full overlap between the "Non-Toxic" and "Ambiguous" classes. This would not be the case if more non-toxic words are added as instances of the ontology.

The ontology consists of two main classes, Content and Context. The Content class explains the categorization of the words in the corpora. The Content class consists of the categories described above, as well as some additional classes that define the restrictions, related to the use of the words in the corresponding contexts. The Context class at other hand is a superclass of the different types of internet contexts.

The main ontology classes that represent the actual content of language units are "Content" and its four subclasses "ToxicLanguage", "MedicalTerminology", "NonToxicLanguage", "MinorityGroup". Each of the subclasses covers the cor- responding categorized words and phrases from the lexicon. Additionally, each of the subclasses contains within it a definition of its category.

In the ontology, we also model the different types of Internet content defined in the Context class, such as:

- **"Forum"** - internet content, typical for discussion forums. Excludes strictly toxic words and phrases (representatives of the "ToxicLanguage" class only), but allows those that are also instances of the "MedicalTerminology", "MinorityGroup", or "NonToxicLanguage" classes. The inclusion of "NonToxicLanguage" allows the use of words that also have a non-toxic meaning. A filter based on the words contained in this class would allow sharing of information related to minorities and reproductive health.
- **"FamilyFriendly"** - content suitable for children. Excludes all words except those with multiple meanings, at least one of which is strictly non-toxic (individuals from the "NonToxicLanguage" classes).

To form a logical definition of context types in the ontology, we first define the sets of words such classes would block as composite subclasses of the Content class. Table 3 represents the class hierarchy and class restrictions described in Description logic (DL). Visualization of the ontology is available in figure 1).

When so defined, context instances will serve as predefined lists of words that can be used for various filters. Modelling in this manner provides a degree of certainty about the context by blocking toxic content, but also preserves a degree of context-specific freedom of expression.

**Table 3.** Logical description of the **ToxicBG** ontology

| Class | Description Logic Representation |
|---|---|
| Context | ⊑ Thing |
| Content | ⊑ Thing |
| ToxicLanguage | ⊑ Content |
| MedicalTerminology | ⊑ Content |
| MinorityGroup | ⊑ Content |
| NonToxicLanguage | ⊑ Content |
| Ambiguous | ≡ ToxicLanguage ⊓ NonToxicLanguage |
| FamilyFriendlyContentBlocked | ≡ (ToxicLanguage ⊔ MedicalTerminology ⊔ MinorityGroup) ⊓ ¬ NonToxicLanguage |
| ForumContentBlocked | ≡ ToxicLanguage ⊓ ¬ NonToxicLanguage ⊓ ¬ MedicalTerminology ⊓ ¬ MinorityGroup |
| FamilyFriendly | ≡ Context ⊓ ∀ blocks.FamilyFriendlyContentBlocked |
| Forum | ≡ Context ⊓ ∀ blocks.ForumContentBlocked |

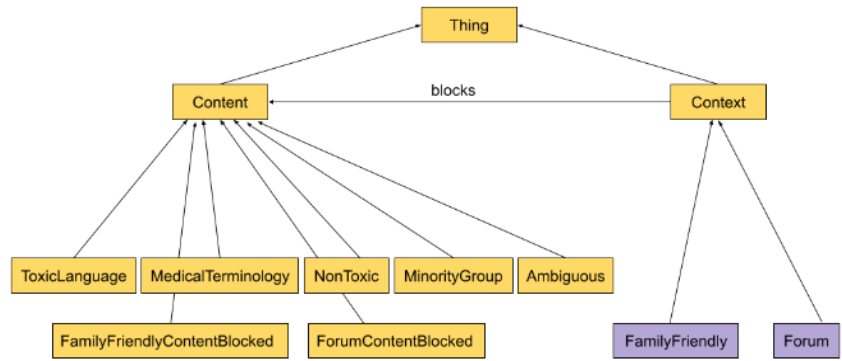

**Fig. 1.** Diagram of the developed ontology for toxic language

### 3.3 Ontology-based Classifier

Building an ontology-based toxic language classifier aims to categorize text ac- cording to the presence of toxic content, medical terminology, non-toxic language, and terms related to minority groups. It uses the developed ontology as a knowledge base that contains categorized words and expressions belonging to these categories.

The classification process starts with the text input to the algorithm. The sentence is split into individual words, and words are cleaned by removing special characters and converting them to lowercase. The algorithm then crawls each word in the sentence and compares it to the individuals in the ontology. If a word in the sentence matches an individual in the ontology, the classifier checks which classes it belongs to and the corresponding categories are marked as present in the text in the internal "classifications" dictionary.

If no match is found, i.e. no word is present in the ontology, the classifier assumes that the sentence does not contain toxic language and marks the category as "non-toxic language" in the internal dictionary. Finally, the function returns the category dictionary, noting that the classified sentence may fall into one or more than one classification. This information can be used to make decisions about further processing or actions related to the text, depending on the specific use case.

The results of the experiment with the ontology-based classifier are summarized in Section 5. Despite the good scores achieved for detecting toxic language, it should be noted that the classifier relies on exact word-to-individual matching in the ontology and does not consider the context or semantics of the whole sentence. This can lead to limitations such as an inability to recognize negation or misclassification of sentences with more complex structure.

Despite the drawbacks, the classifier provides a foundation that can be developed and improved by integrating additional techniques from the fields of natural language processing and machine learning.

## 4      A BERT-based Toxicity Classifier

In order to address some of the limitations of the ontology-based model, we created a machine learning classifier, based on a transformer language model (BERT) - **ToxicBertBg**[5]. To train the model, an existing data corpus[6] was expanded with web-scraped content from Bulgarian online forums. The resulting dataset **ToxicDataBg**[7] comprises 4,384 manually annotated sentences across four categories: toxic language, medical terminology, non-toxic language, and terms related to minority communities. The training process utilized transfer learning on a pre-trained Bulgarian BERT-based model.

### 4.1     Data Preparation

As a starting point, we used the data corpus provided by Ralev and Pfeffer [20], which contains a large number of sentences from online sources in Bulgarian. Initially, we applied an automatic annotation approach using the already developed ontology-based classifier as a tool to pre-annotate the corpus by classifying each sentence into one or more of the defined categories.

The automatic annotation process consists of applying the classifier on each sentence in the corpus. The classification results are recorded in the form of annotations for each sentence. After analyzing the automatic annotation, we noticed that the distribution of categories is highly unbalanced. Out of a total of 106,264 sentences, only 3,044 were classified into any of the categories other than non-toxic language.

To address this imbalance, we scraped 985 additional sentences from the *framar.bg*[8] and 438 from *proud.bg*[9] forums, as well as supplementing with 254 hand-picked non-toxic sentences from the original corpus.

After manual annotation, we obtained a data corpus with 4,384 sentences annotated in one of four categories, which we split into training and test sets in the next step. Figure 2 shows that while an imbalance of the data in favour of the "toxic" ones still exists, a better distribution has been achieved, allowing training a model capable of classifying the four categories.

---

[5]   https://huggingface.co/sofia-uni/toxic-bert-bg

[6]   http://www.pfeffer.at/data/bulgarian/

[7]   https://huggingface.co/datasets/sofia-uni/toxic-data-bg

[8]   The forum forum.framar.bg is for discussions and information related to health, medicine and pharmacy. It includes medical terminology used in a real context.

[9]   The forum proud.bg is aimed at the LQBTQ+ community and discusses topics related to the rights, challenges and experiences of people from this community.

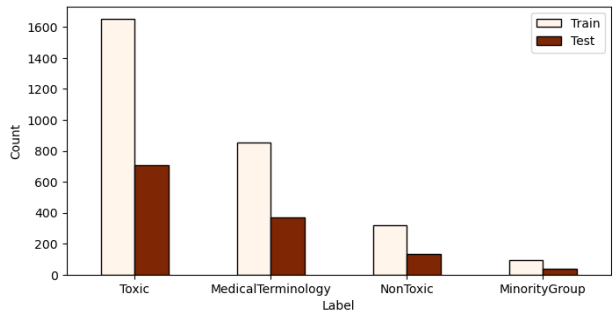

**Fig. 2.** Label distribution of training and test set

### 4.2   Model training

To train a classifier, we used the publicly available model *bert-base-bg*[10], pre-trained on the Bulgarian language corpora of OSCAR, Chitanka, and Wikipedia, which we fine-tuned on our data.

In our model training, we used 5-fold cross-validation to optimize hyperparameters and assess model performance. This approach ensures generalization across diverse data splits, thereby mitigating the risk of overfitting. To facilitate this process, we utilized Optuna[11], a hyperparameter optimization framework that efficiently identified the optimal model configuration. The approach resulted in the following optimal hyperparameter values: number of epochs: 2, learning rate: 5.56e-5, batch size: 32, weight decay: 1.42e-6.

Table 4 presents the averaged cross-validation results across the 5 folds.

**Table 4.** Averaged cross-validation results of **ToxicBertBG**

| Accuracy | Precision | Recall | F1 score | Loss function |
|----------|-----------|--------|----------|---------------|
| 0.85     | 0.86      | 0.85   | 0.85     | 00.43         |

## 5   Evaluation

### 5.1   Comparison of the Classifiers

For the comparison of the two approaches, we used the test set from the data corpus. For the ontology-based classifier, we also use an additional heuristic for toxic class priority to restrict the predicted classifications to only one category and make the estimates comparable. Table 5 shows the classification report for the two models.

---

[10] https://huggingface.co/rmihaylov/bert-base-bg

[11] https://optuna.org/

Table 5. Classification report for the ontology-based and BERT-based classifiers

| Class | Ontology-based classifier | | | BERT-based classifier | | |
|---|---|---|---|---|---|---|
| | Precision | Recall | F1-score | Precision | Recall | F1-score |
| Toxic | 0.94 | 0.54 | 0.69 | 0.92 | 0.91 | 0.92 |
| MedicalTerminology | 0.27 | 0.31 | 0.29 | 0.88 | 0.88 | 0.88 |
| NonToxic | 0.29 | 0.90 | 0.44 | 0.81 | 0.89 | 0.85 |
| MinorityGroup | 0.17 | 0.03 | 0.04 | 0.97 | 0.85 | 0.91 |
| Macro Average | 0.42 | 0.44 | 0.36 | **0.90** | **0.88** | **0.89** |
| Weighted Average | 0.65 | 0.50 | 0.52 | 0.90 | 0.90 | 0.90 |

The classification report shows that the ontology-based approach achieves high precision for toxic language identification, but significantly lower precision for the other classes. On the other hand, the recall values vary dramatically between classes, from 0.90 for non-toxic language to only 0.03 for terms associated with minority communities. The low scores can be explained by the lack of sufficient exemplars from the classes with lower representativeness. A future improvement of the ontology would be to enrich it with words and expressions from the test data. For the BERT-based approach, we observe more balanced results across classes and metrics. The return and precision values are close to each other and are in the range 0.81-0.97. This leads to more uniform and high F1 scores, between 0.85 and 0.92. The overall accuracy of the classifier is 0.90. These results show that the BERT-based classifier outperforms the ontology-based approach in overall effectiveness. However, the ontological method exhibits high precision in identifying specific toxic words and phrases and offers greater flexibility in modelling various contexts.

### 5.2    Comparison With Previous Work

The results we achieved in this paper on the classification of toxic language in a Bulgarian text can be conditionally compared with previous studies on the topic [8,20]. The condition in question is related to the fact that the developed classification models were trained on different corpora and categories and there- fore the results are not directly comparable. The comparison of the F1 macro score of the present model (0.89) with respect to the previous ones shows an improvement – for Ralev and Pfeffer [20] it is 0.72, and for Dinkov et al. [8] it is 0.39.

## 6    Applications for Large Language Models

As large language models (LLMs) are rapidly entering our daily lives, detecting potentially toxic outputs from them is crucial. LMMs are prone to generating text containing toxic language, as they are trained on great amounts of uncurated internet data, such as internet forums.

Various approaches to the problem of toxic language generation exist [25]. This problem is generally perceived as part of the task of model alignment, or achieving model's outputs that are desirable for people. While the task of model alignment is often approached by Reinforcement Learning with Human Feedback (RLHF) and other reinforcement learning methods [21], preventing toxic language generation can also be done with filters built specifically for this task.

As mentioned by Lin et al. [15], identifying the toxic language in language model's outputs can be intricate for current toxicity detection models, as there is a significant difference in the text structure, compared to social media content.

In this section, we will discuss how the proposed approaches can be used for the detection of toxic language and evaluation of toxicity levels in large language models.

### 6.1  Data Preparation

We use the MultiJail dataset [7], consisting of 315 prompts for Jailbreaking[12], marked with tags corresponding to 18 jailbreak categories. Three of the categories fall in the description of toxic language that we use: "Sexual exploitation & human trafficking", "Hate speech & offensive language", and "Adult content".

After selecting the prompts marked with these five categories, we obtain 64 potentially toxic prompts. We then automatically translate the prompts to Bulgarian with the Google Translate API to prepare the prompts for the chosen models.

### 6.2  Models' Responses

We tested 3 models for toxicity generation, the multilingual models Llama 3 by META [9], Gemma 2 by Google [24] and the Bulgarian large language model BgGPT-7B-Instruct-v0.2, based on Mistral architecture [13]. All chosen models are quantized by 4-bit quantization and are publicly available in the Ollama library[13]. We manually annotated the obtained responses into the four categories we used for the classifiers, adhering as closely as possible to the definitions in the ontology. Figure 3 shows the labels' distribution over the tested models' responses. We can see that BgGPT has the highest rate of toxicity, while Gemma 2 has the lowest.

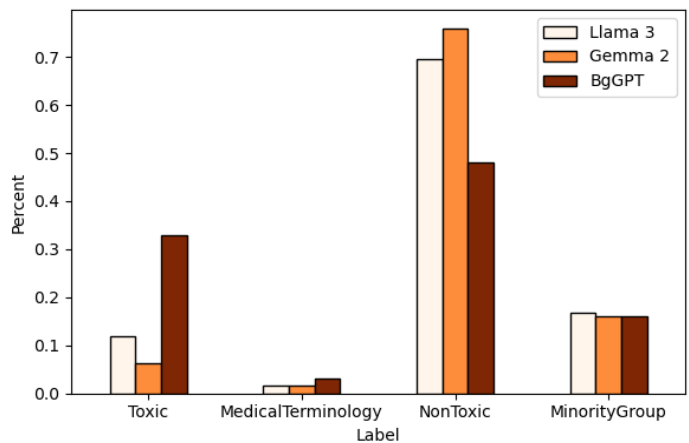

**Fig. 3.** Class distribution of the annotated responses

---

[12] https://huggingface.co/datasets/DAMO-NLP-SG/MultiJail

[13] LLama 3, 8 billion parameters version: https://ollama.com/todorov/bggpt/ blobs/5a2ae4dc1977, Gemma 2, 9 billion parameters version: https://ollama.com/ library/gemma2, BgGPT-7B-Instruct-v0.2: https://ollama.com/todo-rov/bggpt:v0.2

Although all models were fed with Bulgarian-only prompts, the responses of the multilingual ones were not always in Bulgarian. Table 6 shows the language distribution of the model's answers, computed with the Langdetect library[14]. This characteristic of the LLMs complicates toxicity detection, as it becomes a multilingual task. The analysis showed that LLama 3 answers were 90% in English and only 8% in Bulgarian, making it irrelevant to toxicity detection for Bulgarian language. For this reason, we did not include LLama 3 in the classification report (table 7). Gemma 2, on the other hand, responded 85% of the time in Bulgarian and the remaining 15% in Russian and other Slavic languages, and BgGPT responded only in Bulgarian.

Table 6. Language distribution of the models' answers.

| Language  | Llama 3 | Gemma 2 | BgGPT |
|-----------|---------|---------|-------|
| Bulgarian | 0.08    | 0.85    | 1.0   |
| English   | 0.90    | 0.00    | 0.0   |
| Russian   | 0.00    | 0.14    | 0.0   |
| Other     | 0.02    | 0.01    | 0.0   |

### 6.3 Evaluation and Results

To test the success of the classifiers for recognizing toxic content generated by large language patterns, we applied the two presented approaches to the responses of the BgGPT and Gemma 2 models.

We found that the results obtained on the models' responses are significantly lower than those obtained on the test set. This is to a great extent due to the fact that there are large differences in structure, language, and utterance between posts from forums for which these classifiers were tuned, on the one hand, and responses to language models for which we used them out-of-the-box, on the other.

Furthermore, due to the data on which it was trained, the BERT-based model did not learn well enough to recognize language related to human trafficking and adult content. The ontology-based classifier was better able to recognize such content. However, as expected, it fails to identify subtle and indirect forms of harassment and other unwanted content, which results in low recall scores for the classes, different from "NonToxic".

Table 7 shows the classification report for the ontology-based classifier on the preprocessed MultiJail data. These results give us hope that by enriching the data of the ontology, it can be successfully applied in systems for recognizing and filtering toxic content generated by large language models.

Table 7. Classification report for the ontology-based classifier on the BgGPT and Gemma 2 responses.

|                    | BgGPT     |        |          | Gemma 2   |        |          |
|--------------------|-----------|--------|----------|-----------|--------|----------|
| Class              | Precision | Recall | F1-score | Precision | Recall | F1-score |
| Toxic              | 0.71      | 0.21   | 0.32     | 0.0       | 0.0    | 0.0      |
| MedicalTerminology | 1.00      | 0.25   | 0.40     | 0.50      | 0.50   | 0.50     |
| NonToxic           | 0.47      | 0.96   | 0.63     | 0.80      | 0.90   | 0.85     |

---

[14] https://pypi.org/project/langdetect/

| | | | | | | |
|---|---|---|---|---|---|---|
| MinorityGroup | 1.00 | 0.11 | 0.20 | 1.00 | 0.10 | 0.18 |
| Macro Average | 0.80 | 0.38 | 0.39 | 0.58 | 0.38 | 0.38 |
| Weighted Average | 0.67 | 0.52 | 0.44 | 0.80 | 0.73 | 0.71 |

## 7 Conclusion

### 7.1 Contributions

In conclusion, this paper contributes to the topic of toxic language detection in the following directions:

**1. Development of a comprehensive toxic language ontology.** We developed a novel toxic language ontology that effectively differentiates between toxic language, medical terminology, non-toxic language, and terminology related to minority groups. This ontology provides a more nuanced understanding of language usage in various contexts, addressing the issue of toxic content filters mislabeling non-toxic content as toxic. By formally conceptualizing these categories, our ontology enhances the accuracy of content classification.

**2. Creation and annotation of a Bulgarian sentence dataset.** We created and annotated a large dataset of Bulgarian sentences, drawing from various sources such as online forums. This dataset provides a valuable resource for training and evaluating models in the domain of toxic language detection in Bulgarian.

**3. Implementation of ontology-based and BERT-based toxic language classifiers.** We implemented two distinct approaches for toxic content detection in the Bulgarian language. The first approach is based on the developed ontology, allowing for context-sensitive content detection. The second approach uses a model-based method, trained on annotated data, to further refine the classification of toxic content. Both approaches contribute to improving the performance and accuracy of toxic language detection systems.

### 7.2 Limitations

The work described in this paper has some limitations. Firstly, while the annotation process consisted of categorizing the words into one or more ontology classes based on their usage, we did not provide their full definitions, due to the limitations of manual resources. This task will be the subject of future research efforts.

Secondly, the annotation of the Flores lexicon and the train and test corpora can be subjective. While annotating the words in the ontology, we noticed a shift in the perception of the level of toxicity of some words between annotators from different generations. The current annotations reflect mainly the views of the two authors, who are representatives of the so-called "Millennial" and "Gen Z" generations.

The categorization of words that should be blocked in forum and family- friendly contexts is also subjective and can be the subject of further discussion. The demonstrated classifiers also have their limitations. They cannot deal with more complex tasks such as multilingual content and alternative ways of word spelling. The ontology-based classifier is not capable of understanding grammar structures such as negation. Because of that, the classifier will still mark a sentence such as *"As a language model, I could not talk about pornography."* as a toxic one.

### 7.3 Future Work

The research presented in this paper lays a solid foundation for toxic language detection, yet there exist several directions for enhancement and expansion to further augment its efficacy:

**1. Diversification of training data:** To bolster the machine learning algorithm's capacity to detect toxic content in specific domains, such as adult content and human trafficking, it is imperative to incorporate more diverse and representative training data. Expanding the dataset with examples from these areas will enhance the model's ability to recognize and classify such content, which often presents in nuanced and context-dependent forms. The sentences of the annotated forum data can be additionally annotated for the task of toxic span detection.

**2. Ontology enrichment:** While the current ontology provides a robust foundation for toxic language detection, its efficacy can be further enhanced through the addition of more words and phrases. The instances of the ontology can be additionally annotated with their meanings, as well as whether they are part of the official (Bulgarian) language or slang language. This additional labelling can help with the recognition of words that are inappropriate in an official context. Regular updates to the ontology will ensure its continued comprehensiveness and relevance as language evolves and new forms of toxic content emerge.

**3. Multilingual extension:** To improve the detection of toxic content in multilingual contexts, the ontology should be expanded to support multiple languages. This is particularly crucial in online environments where users frequently switch between languages or employ mixed-language expressions. A multilingual ontology will facilitate more accurate detection of toxic con- tent across diverse linguistic contexts, ensuring that harmful language does not evade detection simply due to its expression in a less common or mixed language.

By addressing these areas, future research can significantly enhance the capabilities of toxic language detection systems, rendering them more versatile and effective across a broader spectrum of real-world scenarios. These improvements will contribute substantially to the creation of safer online environments by ensuring that all forms of toxic content are accurately identified and appropriately managed. The implementation of these enhancements will not only refine the current system but also pave the way for more sophisticated and adaptive approaches to content moderation in increasingly complex digital landscapes.

**Acknowledgments.** This study is financed by the European Union-NextGenerationEU, through the National Recovery and Resilience Plan of the Republic of Bulgaria, project №BG-RRP-2.004-0008-C01. Melania Berbatova acknowledges travel support from the European Union's Horizon 2020 research and innovation programme under grant agreement No 951847. We would like to thank Dr. Trifon Trifonov and Dr. Maria Nisheva-Pavlova from Sofia University "St. Kliment Ohridski" and Dr. Erik Derner from ELLIS Aicante Foundation for their kind support and suggestions regarding the paper topic. We are also thankful of Juan José Bayona Reig and Taya Prymak from ELLIS Alicante Foundation for providing a code base for the experiments with large language models that were part of the study.

**Disclosure of Interests.** The authors have no competing interests to declare that are relevant to the content of this article.

Detecting Toxic Language for Bulgarian Text 15